\pdfoutput=1

\documentclass[11pt,a4paper]{article}
\usepackage[table,dvipsnames]{xcolor}
\usepackage[hyperref]{acl2021}

\usepackage[T1]{fontenc}

\usepackage[utf8]{inputenc}

\usepackage{times}
\usepackage{latexsym}

\usepackage{graphicx}               
\usepackage{multirow}               
\usepackage{tabularx}               
\newcolumntype{C}{>{\centering\arraybackslash}X}
\usepackage{amsfonts,amsmath,amssymb,amsthm}
\newcommand{\bs}{\boldsymbol}

\usepackage{microtype}

\usepackage{xparse}

\aclfinalcopy 

\title{Personalized Entity Resolution with Dynamic Heterogeneous Knowledge Graph Representations}

\author{Ying Lin, Han Wang, Jiangning Chen, Tong Wang, Yue Liu, \\
\textbf{Heng Ji, Yang Liu, Premkumar Natarajan}\\
Amazon Alexa AI \\
\texttt{\{linzying,wnghn,cjiangni,tonwng,jihj,yangliud,premknat\}}@amazon.com}
\date{}

\begin{document}

\maketitle

\begin{abstract}
    The growing popularity of Virtual Assistants poses new challenges for Entity Resolution, the task of linking mentions in text to their referent entities in a knowledge base.
    Specifically, in the \textit{shopping} domain, customers tend to use implicit utterances (e.g., ``organic milk'') rather than explicit names, leading to a large number of candidate products.
    Meanwhile, for the same query, different customers may expect different results.
    For example, with ``add \textit{milk} to my cart'', a customer may refer to a certain organic product, while some customers may want to re-order products they regularly purchase.
    To address these issues, we propose a new framework that leverages personalized features to improve the accuracy of product ranking.
    We first build a cross-source heterogeneous knowledge graph from customer purchase history and product knowledge graph to jointly learn customer and product embeddings.
    After that, we incorporate product, customer, and history representations into a neural reranking model to predict which candidate  is most likely to be purchased for a specific customer.
    Experiments show that our model substantially improves the accuracy of the top ranked candidates by 24.6\% compared to the state-of-the-art product search model.

\end{abstract}
\section{Introduction}

Given an entity mention as a query, the goal of entity resolution (or entity linking)~\cite{ji2011knowledge} is to link the mention to its  corresponding entry in a target knowledge base (KB). In an academic shared task setting, an entity mention is usually a name string, which can be a
person, organization or geo-political entity in a news context, and the KB is usually a Wikipedia dump with rich structured properties and unstructured text descriptions. State-of-the-art entity resolution methods can achieve higher than 90\% accuracy in such settings~\cite{ji2011knowledge,Ji2015,agarwal2020entity}, and they have been successfully applied in hundreds of languages~\cite{pan2017crosslingual} and various domains such as disaster management~\cite{Zhang2018} and scientific discovery~\cite{wang2015language}.
Therefore, we tend to think entity resolution is a solved problem in academia.
In industry, with the rise in popularity of Virtual Assistants (VAs) in recent years, an increasing number of consumers now rely on VAs to perform daily tasks involving entities, including shopping, playing music or movies, calling a person, booking a flight,
and managing schedules.
The scale and complexity of industrial applications
presents the following unique new challenges.
\begin{figure*}[!hbt]
    \centering
    \includegraphics[width=\linewidth]{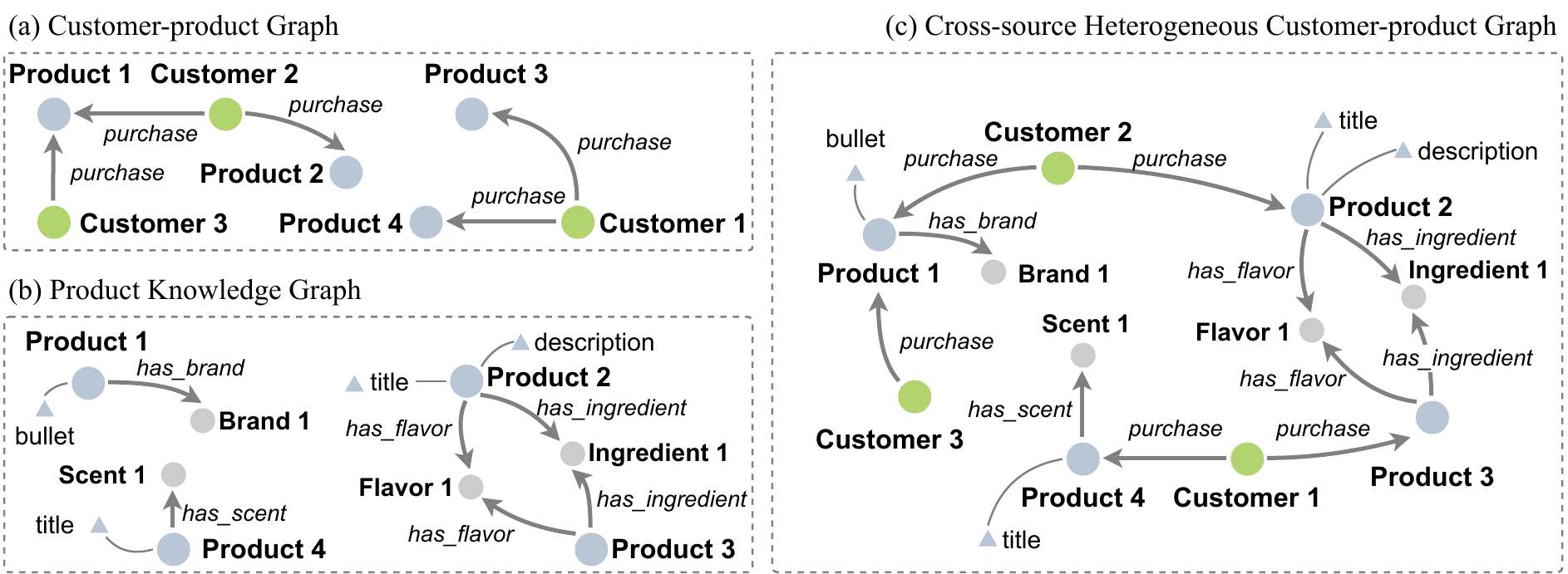}
    \caption{An illustration of the cross-source heterogeneous customer-product graph.}
    \label{fig:enriched_customer_product_graph}
\end{figure*}

\textbf{Unpopular majority.}
There is a massive number of new entities emerging every day. The entity resolver may know very little about them since very few users interact with them. Handling these tail entities effectively requires the use of property linkages between entities and shared user interests

\textbf{Large number of ambiguous variants.}
When interacting with VAs, users tend to use short and less informative utterances with the expectation that the VAs
can intelligently infer their actual intentions.
This further raises the need for the technique to resolve entities
with \textit{personalization}.

In the \textit{shopping} domain, this problem is even more challenging as customers typically use implicit entity reference utterances (e.g., ``organic milk'') instead of explicit names (e.g., ``Horizon Organic Shelf-Stable 1\% Lowfat Milk'')
which usually lead to a large number of candidates due to the ambiguity. However, with VAs' voice user interface (VUI), the number of products that can be shown to the customers is very limited.
In this work, we focus on the problem of \emph{personalized entity resolution} in the shopping domain. Given a query and a list of retrieved candidates, we aim to return the product that is most likely to be purchased by a customer.

We make three assumptions:
(H1) customers tend to purchase products they have purchased in the past;
(H2) customers tend to purchase a set of products that share some properties;
(H3) two customers who purchased products with similar properties may share similar interests.
Based on these assumptions, we propose to represent customers and products as low-dimensional distributional vectors learned from a graph of customers and products.
However, unlike social media sites with rich interactions
among users, customers of most shopping services are isolated,
which prevents us from learning user embeddings as distributed representations.
To address this issue,
we propose to build a cross-source heterogeneous knowledge graph as Figure~\ref{fig:enriched_customer_product_graph} depicts to establish rich connections among customers from two data sources, users' purchase history (customer-product graph) and product knowledge graph,  and further jointly learn the representations of nodes in this graph using a Graph Neural Network (GNN)-based method.
We further propose an attentive model to generate a query-aware history representation for each user based on the current query.

Experiments on real data collected from an online shopping service
show that our method substantially improves the purchase rate and revenue of the top ranked products.

\section{Methodology}

\begin{figure}[!hbt]
    \centering
    \includegraphics[width=\linewidth]{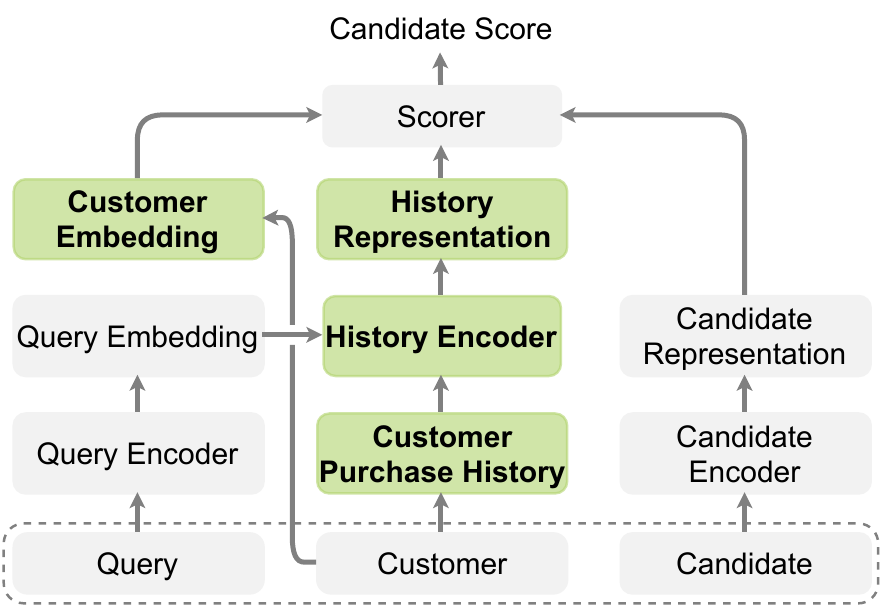}
    \caption{An illustration of our framework.
    }
    \label{fig:overview}
\end{figure}

Given a query $q$ from a customer $c$, and a list of candidate products $\mathcal{P} = \{p_1, ..., p_L\}$, where $L$ is the number of candidates, our goal is to predict the product that the customer will shop for based on the customer's purchase history and the product knowledge graph.
Specifically, we use purchase records $\{r_1, ..., r_H\}$ where $H$ is the number of historical records.
As Figure~\ref{fig:overview} illustrates,
we jointly learn customer and product embeddings from a cross-source customer-product graph using GNN.
To perform personalized ranking, we incorporate the learned customer embedding and the query-aware history representation as additional features when calculating the score of each candidate.
We then rank all candidates by score and return the top one.

\subsection{Candidate Retrieval}

We first retrieve candidate products for each query using QUARTS~\cite{nigam2019productsearch,nguyen-etal-2020-learning}, which is an end-to-end neural model for product search. QUARTS has three major components: (1) an LSTM-based (long short-term memory) classifier that predicts whether a query-product pair is matched;
(2) a variational
query generator that generates difficult negative examples, and (3) a state combiner that switches between query representations computed by the classifier and generator.

\subsection{Joint Customer and Product Embedding}

The next step is to obtain the representations of customers and products.
Customer embeddings are usually learned from user-generated texts~\cite{preoctiuc2015analysis,yu2016user,Ribeiro2018CharacterizingAD} or social relations~\cite{Perozzi2014DeepWalkOL,grover2016node2vec,zhang2018anrl}, neither of which are available in the shopping dataset we use.
Alternatively, we establish \textit{indirect} connections among customers through their purchased products under hypothesis H3, and form a customer-product graph as shown in Figure~\ref{fig:enriched_customer_product_graph}(a).
This graph only contains a single type of relation (i.e., \texttt{purchase}) and ignores product attributes. As a result, it tends to be sparse and less effective for customer representation learning.

In order to learn more informative embeddings, we propose to incorporate richer information from a product knowledge graph
(Figure~\ref{fig:enriched_customer_product_graph}(b)) where products are not only connected to different attribute nodes (e.g., brands, flavors),
but they may also be associated with textual features (e.g., \texttt{title}) and boolean features (e.g., \texttt{isOrganic}).

By merging the product knowledge graph and the customer-product graph, we obtain a more comprehensive graph (Figure~\ref{fig:enriched_customer_product_graph}(c)) of higher connectivity.
For example, in the original customer-product graph, \texttt{Customer 1} and \texttt{Customer 2} are disconnected because they do not share any purchase.
In the new graph, they have an indirect connection through \texttt{Product 2} and \texttt{Product 3}, which share the same flavor and ingredient.

From this heterogeneous graph, we jointly learn customer and product representations using a two-layer Relational Graph Convolutional Network~\cite{schlichtkrull2018modeling}.
The embedding of each node is updated as:

\begin{equation*}
    \bs{h}^{l+1}_i = \mathrm{ReLU}\Big( \bs{W}^l_0 \bs{h}^l_i + \sum_{r \in R}\sum_{j \in N_i^r}\frac{1}{|N_i^r|}{\bs{W}^l_r \bs{h}^l_j} \Big),
\end{equation*}
where $\bs{h}_i^l$ is the representation of node $i$ at the $l$-th layer, $N_i^r$ is the set of neighbor indices of node $i$ under relation $r \in R$, and $\bs{W}^l_0$ and $\bs{W}^l_r$ are learnable weight matrices.

In order to capture textual features such as product titles and descriptions, we use a pre-trained RoBERTa~\cite{liu2019roberta} encoder to generate a fix-sized representation for each product. Specifically, we concatenate textual features using a special separator token \texttt{[SEP]},
obtain the RoBERTa representation for each token, and then use the averaged embedding to represent the whole sequence.
To reduce the runtime, we calculate customer and product embeddings offline and cache the results.

\subsection{Candidate Representation}

In addition to the product embedding, we further incorporate the following features to enrich the representation of each candidate.

\textbf{Rank}: the order of the candidate returned by the product retrieval system.

\textbf{Relative Price}: how much a product's absolute price is higher or lower than the average price of all retrieved candidates.

\textbf{Previously Purchased}: a binary flag indicating whether a candidate has been purchased by the customer or not.

We concatenate these features with the product embedding and project the vector into a lower dimensional space using a feed forward network.

\subsection{History Representation}
\label{sec:history_representation}

Although customer embeddings can encode purchase history information,
they are static and may not effectively provide the most relevant information for each specific query.
For example, if the query is ``bookshelf'', furniture-related purchase records are more likely to help the model predict the product that the customer will purchase, while if the query is ``sulfate-free shampoo'', purchase records of beauty products are more relevant.
To tackle this issue, we propose to generate a dynamic history representation $\boldsymbol{v}$ based on the current query $\bs{q}$ from all purchase record representations $\{\boldsymbol{v}_1, ..., \boldsymbol{v}_H\}$ of the customer.

We first represent each purchase record as the concatenation of the product embedding, product price, and purchase timestamp.
The query-aware history representation is then calculated as a weighted sum of the customer's purchase record representations using an attention mechanism as follows.

\begin{equation*}
    e_i = \bs{v}^\top \tanh\big( \bs{W}_q \bs{q} + \boldsymbol{W}_v \bs{v}_i \big),
\end{equation*}
\begin{equation*}
    a_i = \mathrm{Softmax}(e_i) = \frac{\exp{(e_i)}}{\sum_k^M \exp{(e_k)}},
\end{equation*}
\begin{equation*}
    \bs{v} = \sum_i^H{a_i\bs{v}_i}.
\end{equation*}

\subsection{Candidate Ranking}

We adopt a feed forward neural network
that takes in the candidate, customer, and history representations, and returns a confidence score $\hat{y}_i$.
The confidence score is scaled to $(0,1)$ using a Sigmoid function.
During training, we optimize the model by minimizing the following binary cross entropy loss function.
\begin{equation*}
    \mathcal{L} = -\frac{1}{N} \sum_{i=1}^N{y_i \log{\hat{y}_i} + (1 - y_i)\log{(1 - \hat{y}_i)}},
\end{equation*}
where $N$ denotes the total number of candidates, and $y_i \in \{0, 1\}$ is the true label.
In the inference phase, we calculate confidence scores for all candidates for each session and return the one with the highest score.
\section{Experiment}

\subsection{Data}

\noindent\textbf{Product Knowledge Graph}.
In our experiment, we use a knowledge graph of products in five categories (i.e., grocery, beauty, luxury beauty, baby, and health care), which contains 24,287,337 unique product entities.
As Figure~\ref{fig:enriched_customer_product_graph} depicts, products in this knowledge graph are connected through attribute nodes, including brands, scents, flavors, and ingredients.
This knowledge graph also provides rich attributes for each product node.
We use two types of attributes in this work, textual features (i.e., title, description, and bullet)
and binary features (e.g., \texttt{isOrganic}, \texttt{isNatural}).

\noindent\textbf{Evalution Dataset}.
We randomly collect 1 million users' purchase sessions from November 2018 to October 2019 on an online shopping service.
Each session contains a query, an obfuscated identifier, a timestamp, and a list of retrieved candidate products
where only one product is purchased.

We split sessions before and after September 1, 2019 into two subsets.
The first subset only serves as the purchase history and is used to construct the customer-product graph.
From the second subset, we randomly sample 22,000 customers with at least one purchase record in the first subset and take their last purchase sessions for training or evaluation.
Specifically, we use 20,000 sessions for training, 1,000 for validation, and 1,000 for test.
If a customer has multiple purchase sessions in the second subset, other sessions before the last one are also considered as purchase history when we generate history representations, while they are excluded from the customer-product graph, which is constructed from the first subset.

\subsection{Experimental Setup}

We optimize our model with AdamW for 10 epochs with a learning rate of 1e-5 for the RoBERTa encoder, a learning rate of 1e-4 for other parameters, weight decay of 1e-3, a warmup rate of 10\%, and a batch size of 100.

To encode textual features, we use the RoBERTa base model\footnote{\url{https://huggingface.co/transformers/pretrained_models.html}} with an output dropout rate of 0.5.
To represent query words, we use 100-dimensional GloVe embeddings pre-trained on Wikipedia and Gigaword\footnote{\url{https://nlp.stanford.edu/projects/glove/}}.
We set the size of pre-trained customer and product embeddings to 100 and freeze them during training.

We use separate fully connected layers to project candidate and history representations into 100-dimensional feature vectors before concatenating them for ranking.
We use a two-layer feed forward neural network with a hidden layer size of 50 as the ranker and apply a dropout layer with a dropout rate of 0.5 to its input.

\subsection{Quantitative Analysis}

We compare our model to the state-of-the-art
product search model QUARTS as the baseline.
Because our target usage scenarios are VAs where only one result will be returned to the user, we use Accuracy@1 as our evaluation metric.
We implement the following baseline ranking methods.

\noindent\textbf{Purchased}: We prioritize products previously purchased by the customer. If multiple candidates are previsouly purchased, we return the one ranked higher by QUARTS.

\noindent\textbf{ComplEx}: Customer and product embeddings are learned using ComplEx~\cite{trouillon2016complex}, a widely used knowledge embedding model.

In Table~\ref{table:quantitative_result}, we show the relative gains compared to the baseline model QUARTS.
With personalized features, our method effectively improves Acc@1 on both development and test sets.

We also conduct ablation studies by removing the following features and show results in Table~\ref{table:ablation_study}.

\noindent\textbf{Ranking}: In this setting, our model ignores the original retrieval ranking returned by QUARTS.

\noindent\textbf{Personalized Features}:
We remove personalized features (e.g., customer embedding, whether a product is previously purchased) in this setting.

\noindent\textbf{Product Embedding}:
We remove pre-trained product embedding but still use textual features and binary features to represent products.

\noindent\textbf{Joint Embedding}:
Customer and product embeddings are not jointly learned from the merged graph.
Alternatively, customer embeddings are learned from the customer-product graph, and product embeddings are learned from the product knowledge graph.

In Table~\ref{table:ablation_study}, from the results of Methods 6 and 7, we can see that removing either product or customer embedding degrades the performance of the model.
The result of Method 8 shows that embeddings jointly learned from the merged cross-source graph achieve better performance on our downstream task.
We also observe that the ranking returned by the product search system is still an important feature as Method 6 shows.

\begin{table}[!hbt]
    \small
    \centering
    \setlength\tabcolsep{3pt}
    \setlength\extrarowheight{1pt}
    \begin{tabularx}{\linewidth}{|l|X|C|C|}
    \hline

    & \textbf{Method} & \textbf{Dev Acc@1} & \textbf{Test Acc@1} \\
    \hline

    1 & QUARTS & 0.0 & 0.0 \\
    2 & Purchased & +10.5 & +8.5 \\
    3 & ComplEx & +25.7 & +16.1 \\
    4 & Our Model & \textbf{+32.9} & \textbf{+24.6}\\

    \hline
    \end{tabularx}
    \caption{Relative gains compared to QUARTS. (\%)}
    \label{table:quantitative_result}
\end{table}
\begin{table}[!hbt]
    \small
    \centering
    \setlength\tabcolsep{3pt}
    \setlength\extrarowheight{1pt}
    \begin{tabularx}{\linewidth}{|l|lCC|}
    \hline

    & \textbf{Method} & \textbf{Dev Acc@1} & \textbf{Test Acc@1} \\
    \hline

    4 & Our Model & \textbf{+32.9}  & \textbf{+24.6} \\
    5 & \ \ w/o Ranking & -17.1 & -20.4  \\
    6 & \ \ w/o Personalized Features & -10.5 & -18.0 \\
    7 & \ \ w/o Product Embedding & +25.2 & +19.0 \\
    8 & \ \ w/o Joint Embedding & +28.1 & +20.4 \\
    \hline
    \end{tabularx}
    \caption{Ablation study. (\%, relative gains compared to QUARTS.)}
    \label{table:ablation_study}
\end{table}

\subsection{Qualitative Analysis}

\newcolumntype{S}{>{\raggedright\hsize=.36\hsize}X}
\newcolumntype{M}{>{\hsize=1.32\hsize}X}
\newcolumntype{L}{>{\hsize=2.64\hsize}X}

\begin{table*}[!hbt]
    \small
    \centering
    \setlength\tabcolsep{3pt}
    \setlength\extrarowheight{3pt}
    \begin{tabularx}{\linewidth}{|S|M|M|}
    \hline
    \textbf{Query} & \textbf{Candidates} & \textbf{History} \\
    \hline

    \#1 vitamin c serum & \cellcolor{LimeGreen!20}* [3] instanatural vitamin c serum with hyaluronic acid \& vit e - natural \& organic anti wrinkle ... & * \textcolor{black!60}{foundation makeup brush flat top kabuki for face - perfect for blending liquid, cream or flawless powder} \\
    & * [1] truskin vitamin c serum for face, topical facial serum with hyaluronic acid, vitamin e, 1 fl oz & * \textcolor{black!60}{women's rogaine 5\% minoxidil foam for hair thinning and loss, topical treatment for women's hair ...} \\
    & * [2] vitamin c serum for face - anti aging facial serum & * \textcolor{black!60}{vita liberata advanced organics fabulous self-tanning gradual lotion with marula oil, 6.76 fl oz} \\
    & * [4] vitamin c serum plus 2\% retinol, 3.5\% niacinamide, 5\% hyaluronic acid, 2\% salicylic acid ... & * \cellcolor{CadetBlue!10}instanatural vitamin c serum with hyaluronic acid \& vit e - natural \& organic anti wrinkle reducer ... \\
    \cline{2-3}
    &\multicolumn{2}{l|}{Our model promotes candidate 3 as this product was purchased by the customer.}\\
    \hline

    \#2 toothpaste & * \cellcolor{LimeGreen!20}[2] crest 3d white whitening toothpaste, radiant mint, 3.5oz, twin pack & * \cellcolor{CadetBlue!10}crest 3d white toothpaste radiant mint (3 count of 4.1 oz tubes), 12.3 oz packaging may vary \\
    & * \textcolor{black!60}{[1] crest + scope complete whitening toothpaste, minty fresh, 5.4 oz, pack of} 3 & * \textcolor{black!60}{skindinavia the makeup of countrol finishing spray, 8 fluid ounce}\\
    & * \textcolor{black!60}{[3] pronamel gentle whitening enamel toothpaste for sensitive teeth, alpine breeze-4 ounces (pack of 3)} & * \cellcolor{CadetBlue!10}crest 3d white toothpaste radiant mint (3 count of 4.1 oz tubes), 12.3 oz packaging may vary \\
    & * \textcolor{black!60}{[4] colgate cavity protection toothpaste with fluoride - 6 ounce (pack of 6)} & * \textcolor{black!60}{nivea shea daily mointure body lotion - 48 hour moisture for dry skin - 16.9 fl. oz. pump bottle, ...} \\
    \cline{2-3}
    & \multicolumn{2}{L|}{Although the previously purchased item is no longer available, with entity embedding learned from the cross-source graph, our model successfully promotes the most similar product.}\\
    \hline
    \#3 sun dried tomatoes & * \cellcolor{LimeGreen!20}[3] 365 everyday value, \textbf{organic} sundried tomatoes in extra virgin olive oil, 8.5 oz & \cellcolor{CadetBlue!10}* \#1 usda \textbf{organic} aloe vera gel - no preservatives, no alcohol - from freshly cut usa grown 100\% pure ... \\
    & * \textcolor{black!60}{[1] 35 oz bella sun luci sun dried tomatoes julienne cut in olive oil (original version)} & \cellcolor{CadetBlue!10}* \textbf{organic} aloe vera gel with 100\% pure aloe from freshly cut aloe plant, not powder - no xanthan ...\\
    & * \textcolor{black!60}{[2] julienne sun-dried tomatoes - 16oz bag (kosher)} & \cellcolor{CadetBlue!10}* wicked joe \textbf{organic} coffee wicked italian ground \\
    & * \textcolor{black!60}{[4] organic sun-dried tomatoes with sea salt, 8 ounces - salted, non-gmo, kosher, raw, vegan, ...} & *\textcolor{black!60}{thayers alcohol-free original witch hazel facial toner with aloe vera formula, clear, 12oz} \\
    \cline{2-3}
    & \multicolumn{2}{L|}{Our model promotes an organic product as the customer probably prefers organic products based on the shopping records.}\\
    \hline
    \end{tabularx}
    \caption{Positive examples in the test set.
    Candidates are listed in the order returned by our method.
    The number before each candidate is the original ranking returned by QUARTS.
    In the candidate column, we highlight the \colorbox{LimeGreen!20}{purchased products}. In the history column, we highlight \colorbox{CadetBlue!10}{related records}.
    }
    \label{table:positive_example}
    \end{table*}

In Table~\ref{table:positive_example} and Table~\ref{table:negative_example}, we show some positive and negative examples in the test set.

\begin{table*}[!hbt]
    \small
    \centering
    \setlength\tabcolsep{3pt}
    \setlength\extrarowheight{3pt}
    \begin{tabularx}{\linewidth}{|S|M|M|}
    \hline
    \textbf{Query} & \textbf{Candidates} & \textbf{History} \\
    \hline

    \#4 wasabi almonds & * \textcolor{black!60}{[8] blue diamond almonds, bold wasabi \& soy sauce, 16 ounce (pack of 1)} & * \textcolor{black!60}{epsoak epsom salt 19 lb. bulk bag magnesium sulfate usp}\\
    & * \textcolor{black!60}{[2] blue diamond almonds variety pack (1.5 ounce bags) (20 pack)} & \cellcolor{CadetBlue!10}* blue diamond almonds, bold wasabi \& soy sauce, 16 ounce (pack of 1)\\
    & * \textcolor{black!60}{[1] blue diamond almonds bold wasabi \& soy sauce almonds, 25 ounce (pack of 1)} & * \textcolor{black!60}{signature trail mix, peanuts, m \& m candies, raisins, almonds \& cashews, 4 lb}\\
    & * \cellcolor{LimeGreen!20}[6] blue diamond almonds, bold wasabi \& soy, 1.5 ounce (pack of 12) & * \textcolor{black!60}{amazon brand - happy belly nuts, chocolate \& dried fruit trail mix, 48 ounce} \\
    \cline{2-3}
    & \multicolumn{2}{L|}{Our model promotes candidate 8 which is previously purchased, whereas the customer selects another size.}\\
    \hline

    \#5 cacao powder & * \textcolor{black!60}{[5] anthony's organic cocoa powder, 2 lb, batch tested and verified gluten free \& non gmo} & \cellcolor{CadetBlue!10}* anthony's organic cocoa powder, 2 lb, batch tested and verified gluten free \& non gmo\\
    & * \textcolor{black!60}{[1] viva naturals \#1 best selling certified organic cacao powder from superior criollo beans, 1 lb bag} & * \textcolor{black!60}{v\"or all natural keto nut butter spread (10oz) | only two ingredients | no sugar, no salt | vegan ...} \\
    & * \textcolor{black!60}{[2] navitas organics cacao powder, 16oz. bag - organic, non-gmo, fair trade, gluten-free} & \cellcolor{CadetBlue!10}* anthony's organic cocoa powder, 2 lb, batch tested and verified gluten free \& non gmo\\
    & * \textcolor{black!60}{[3] terrasoul superfoods raw organic cacao powder, 1 lb - raw | keto | vegan} & * \textcolor{black!60}{nutiva organic, neutral tasting, steam refined coconut oil from non-gmo, sustainably farmed coconuts ...}\\
    & * \cellcolor{LimeGreen!20}[4] viva naturals certified organic cacao powder (2lb) for smoothie, coffee and drink mixes & \\
    \cline{2-3}
    & \multicolumn{2}{L|}{Our model promote ``Anthony's Organic Cocoa Powder'' as it has been purchased twice by the customer.}\\
    \hline

    \end{tabularx}
    \caption{Negative examples in the test set.}
    \label{table:negative_example}
\end{table*}
Table~\ref{table:negative_example} shows examples where our model fails to return the correct item.
In many cases, such as Example \#4, the purchased product and the top ranked one only differ in packaging size.
We also observe that sometimes customers may not repurchase a product even if it is in the candidate list.

\begin{figure}[!hbt]
    \centering
    \includegraphics[width=\linewidth]{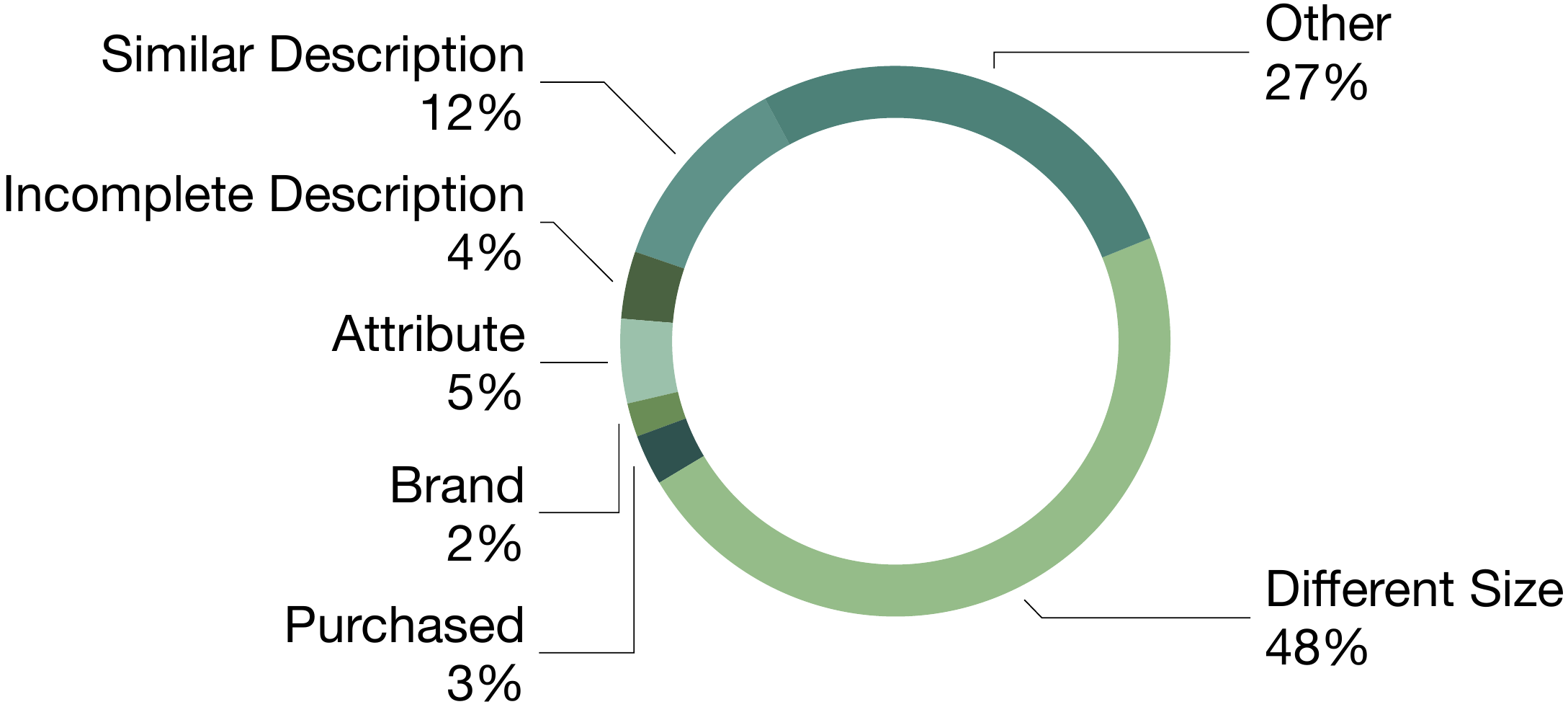}
    \caption{Distribution of remaining Errors.
    }
    \label{fig:error_pie}
\end{figure}

To better understand the remaining errors, we randomly sample 100 examples where our model fails to predict the purchased items.
As Figure~\ref{fig:error_pie} illustrates, we analyze these examples and classify the possible reasons into the following categories.

\noindent\textbf{Different size}.
The predicted product and ground truth are the same product but differ in size.
For example, while our model predicts ``Lipton Herbal Tea Bags, Peach Mango, 20 ct'', the customer purchases another item ``Lipton Tea Herbal Peach Mango (pack of 2)'', which is actually the same product in 2 pack.

\noindent\textbf{Purchased}. The customer has purchased the predicted product but decides not to repurchase it.
This usually happens in categories (e.g., toothpaste) where customers are more willing to try new products.
Additionally, customers may be less likely to repurchase a product in some categories such as books and electronics.

\noindent\textbf{Uninformative title}. The purchased product has an uninformative title and is therefore not promoted.
For example, when the customer searches for ``masaman curry paste maesri'', our model promotes ``Maesri Thai Masaman Curry - 4 oz (pack of 4)'', while the customer purchases ``6 Can (4oz. Each) of Thai Green Red Yellow Curry Pastes Set'', which is also a Maesri product, but this key information is missing from its title.

\noindent\textbf{Similar title}. The title of the predicted product is similar to the titles of some purchased products in the customer's history in a less important aspect.
For example, the model promotes a ``moisturizing'' shave gel because the customer has purchased a ``moisturizing'' body wash, whereas the customer decides to purchase a product for ``sensitive skin''.

\noindent\textbf{Brand}. The customer has purchased one or more products of the same brand.

\noindent\textbf{Attribute}. The customer has purchased one or more products with the same attribute (e.g., organic, keto, kosher).

\noindent\textbf{Other}.
The model may fail to predict the purchased item in other uncategorized cases.
For example, when a customer searches for ``nail clippers'' but has purchased only food in the past, the model is unlikely to utilize the history records to improve the ranking.

Although our framework can improve the accuracy of predicting products that will be purchased, there are still some remaining challenges.

\noindent\textbf{Incorporating more informative features}.
Some important features that affect purchase decisions are still missing in our framework, such as the average rating, customer reviews, and number of ratings.
For example, we may promote the highest rated product for a customer who usually buys products with high ratings.

\noindent\textbf{Building a more comprehensive cross-source customer-product graph}.
In this work, we merge the customer-product graph and product knowledge graph into a single graph, which has been proved to produce better embeddings for our target task.
A natural extension is to include records from more sources, such as music or video playing history, and multimedia features.

\noindent\textbf{Modeling the interactions among purchase behaviors}.
Our current attention-based method that generates history representations is ``flat'' and ignores the relationship among purchase behaviors.
For example, for a customer who previously purchases a pod coffee maker, we should promote coffee capsules in the candidates over coffee beans or grounds.

\section{Related Work}

\subsection{Neural Entity Linking}
A variety of neural models~\cite{gupta2017entity,kolitsas2018end,cao2018neural,sil2018neural,gillick2019learning,logeswaran2019zero,wu2019scalable,agarwal2020entity} have been applied to Entity Linking in recent years.
Compared to traditional entity linking, our task is different in three aspects:
(1) Our mentions are typically vague and occur in uninformative contexts, such as ``add \textit{toothpaste} to my cart'' ;
(2) A mention may be reasonably linked to multiple entities, while only one of them is considered ``correct'' (purchased by the customer);
(3) The ground truth for the same mention can be different for different customers.

\subsection{Personalized Recommendation}

A recommender system is an information filtering system that aims to suggest a list of items in which a user may be interested.
Content-based filtering~\cite{billsus2000user,aciar2007informed,wang2018content} and collaborative filtering~\cite{shardanand1995social,konstan1997grouplens,linden2003amazon,zhao2010user} are two common approaches used in recommender systems.
In recent years, researchers have also applied neural methods to improve the quality of recommendations~\cite{xue2017deep,he2017neural,wang2019knowledgeaware,wang2019multi}.
Recommender systems usually rank items based on the user's past behaviors (e.g., purchasing, browsing, rating) and current context \cite{linden2003amazon,smith2017two}, whereas the results are not constrained by queries.
Instead, our task requires a specific query and only returns the product that is most likely to be purchased from a list of relevant candidates.

\subsection{Graph Embedding}

Various methods have been proposed to learn low-dimensional vectors for nodes in knowledge graphs.
Knowledge graph embedding methods, such as TransE~\cite{bordes2013translating}, DistMult~\cite{yang2014embedding}, ComplEx~\cite{trouillon2016complex}, and RotatE~\cite{sun2018rotate}, typically represent the head entity, relation, and tail entity in each triplet in the knowledge graph as vectors and aim to rank true triplets higher than corresponding corrupted triplets.
Matrix Factorization-based methods~\cite{he2004locality,nickel2011three,qiu2018network} represent the graph as a matrix and obtain node vectors by factorizing this matrix.
Another category of frameworks~\cite{perozzi2014deepwalk,yang2015multi,grover2016node2vec} use random walk to sample paths from the input graph and learn node embeddings from the sampled paths using neural models such as SkipGram and LSTM.

\section{Conclusion and Future Work}
We propose a novel framework to jointly learn customer and product representations based on a cross-source heterogeneous graph constructed from customers' purchase history and the product knowledge graph to improve personalized entity resolution.
Experiments show that our framework can effectively increase the purchase rate of top ranked products.
In the future, we plan to investigate better approaches to integrating personalized features and extend the framework to cross-lingual cross-media settings and generate conversations for more pro-active and explainable entity recommendation and summarization.

\bibliography{acl2021}
\bibliographystyle{acl_natbib}

\end{document}